\documentclass[acmlarge]{acmart}

\makeatletter
\newcommand{\myconfshort}{\acmConference@shortname}
\newcommand{\myconffull}{\acmConference@name}
\newcommand{\myconfdate}{\acmConference@date}
\newcommand{\myconfloc}{\acmConference@venue}
\AtBeginDocument{
  \fancypagestyle{firstpagestyle}{
    \fancyhead{}%
    \fancyfoot[C]{}%
  }
  \fancyhf{}
  \fancyhead[LO]{\@headfootfont\shorttitle}%
  \fancyhead[RE]{\@headfootfont\@shortauthors}%
  \fancyhead[LE]{\@headfootfont\footnotesize \myconfshort, \myconfdate, \myconfloc}%
  \fancyhead[RO]{\@headfootfont\footnotesize \myconfshort, \myconfdate, \myconfloc}%
  \fancyfoot[C]{}%
}
\makeatother
\acmBooktitle{\conffull\@ (\confshort), \confdate, \confloc}

\makeatletter
\def\@ACM@checkaffil{
    \if@ACM@instpresent\else
    \ClassWarningNoLine{\@classname}{No institution present for an affiliation}%
    \fi
    \if@ACM@citypresent\else
    \ClassWarningNoLine{\@classname}{No city present for an affiliation}%
    \fi
    \if@ACM@countrypresent\else
        \ClassWarningNoLine{\@classname}{No country present for an affiliation}%
    \fi
}
\makeatother

\usepackage{booktabs} 
\usepackage{amsmath}
\usepackage{xcolor}
\usepackage{graphicx}
\usepackage{tabularx}
\usepackage{multicol}
\usepackage{xltabular}

\setcopyright{none}             
\renewcommand\footnotetextcopyrightpermission[1]{} 

\newif\ifshowcomments
\showcommentstrue 

\ifshowcomments
    \newcommand{\sheza}[1]{\textcolor{purple}{[Sheza: #1]}}
    \newcommand{\ishtiaque}[1]{\textcolor{blue}{[Ishtiaque: #1]}}
    \newcommand{\krisha}[1]{\textcolor{orange}{[Krisha: #1]}}
    \newcommand{\ben}[1]{\textcolor{brown}{[Ben: #1]}}
    \newcommand{\julian}[1]{\textcolor{red}{[Julian: #1]}}
    \newcommand{\shivani}[1]{\textcolor{violet}{[Shivani: #1]}}
    \newcommand{\edith}[1]{\textcolor{blue}{[Edith: #1]}}
    \newcommand{\ram}[1]{\textcolor{purple}{[Ram: #1]}}
    \newcommand{\ding}[1]{\textcolor{orange}{[Ding: #1]}}
\else
    \newcommand{\sheza}[1]{}
    \newcommand{\ishtiaque}[1]{}
    \newcommand{\krisha}[1]{}
    \newcommand{\ben}[1]{}
    \newcommand{\julian}[1]{}
    \newcommand{\shivani}[1]{}
    \newcommand{\edith}[1]{}
    \newcommand{\ram}[1]{}
    \newcommand{\ding}[1]{}
\fi

\author{Sheza Munir}
\affiliation{%
 \institution{University of Toronto}
 \department{Computer Science}}
\email{sheza@cs.toronto.edu}

 \author{Benjamin Mah}
\affiliation{%
 \institution{University of Toronto}
 \department{Engineering Science}}
\email{benjamin.mah@mail.utoronto.ca}

 \author{Krisha Kalsi}
\affiliation{%
 \institution{University of Toronto}
 \department{Computer Science}}
\email{krisha.kalsi@mail.utoronto.ca}

  \author{Shivani Kapania}
 \affiliation{%
 \institution{Carnegie Mellon University}
 \department{School of Computer Science}}
\email{skapania@andrew.cmu.edu}
 
  \author{Julian Posada}
 \affiliation{%
 \institution{Yale University}
 \department{American Studies}}
\email{julian.posada@yale.edu}
 
 \author{Edith Law}
 \affiliation{%
 \institution{University of Waterloo}
 \department{Computer Science}}
\email{edithlaw@uwaterloo.ca}
 
  \author{Ding Wang}
\affiliation{%
 \institution{Google Research}}
\email{drdw@google.com}
 
 \author{Syed Ishtiaque Ahmed}
\affiliation{%
 \institution{University of Toronto}
 \department{Computer Science}}
\email{ishtiaque@cs.toronto.edu}

\copyrightyear{2026}
\acmYear{2026}
\setcopyright{cc}
\setcctype{by}
\acmConference[FAccT '26]{The 2026 ACM Conference on Fairness, Accountability, and Transparency}{June 25--28, 2026}{Montreal, QC, Canada}
\acmBooktitle{The 2026 ACM Conference on Fairness, Accountability, and Transparency (FAccT '26), June 25--28, 2026, Montreal, QC, Canada}
\acmDOI{10.1145/3805689.3806510}
\acmISBN{979-8-4007-2596-8/2026/06}

\renewcommand{\shortauthors}{Munir, et al.}

\begin{document}


\title[The Consensus Trap]{The Consensus Trap: Dissecting Subjectivity and the "Ground Truth" Illusion in Data Annotation}

\begin{abstract}

In machine learning, "ground truth" refers to the assumed correct labels used to train and evaluate models. However, the foundational "ground truth" paradigm rests on a positivistic fallacy that treats human disagreement as technical noise rather than a vital sociotechnical signal. This systematic literature review analyzes research published between 2020 and 2025 across seven premier venues: ACL, AIES, CHI, CSCW, EAAMO, FAccT, and NeurIPS, investigating the mechanisms in data annotation practices that facilitate this "consensus trap". Our reflexive thematic analysis of $346$ papers reveals that systemic failures in positional legibility, combined with the recent architectural shift toward human-as-verifier models, specifically the reliance on model-mediated annotations, introduce deep-seated anchoring bias and effectively remove human voices from the loop. We further demonstrate how geographic hegemony imposes Western norms as universal benchmarks, often enforced by the performative alignment of precarious data workers who prioritize requester compliance over honest subjectivity to avoid economic penalties. Critiquing the "noisy sensor" fallacy, where statistical models misdiagnose pluralism as error, we argue for reclaiming disagreement as a high-fidelity signal essential for building culturally competent models. To address these systemic tensions, we propose a roadmap for pluralistic annotation infrastructures that shift the objective from discovering a singular "right" answer to mapping the diversity of human experience.
\end{abstract}



\maketitle 
\makeatletter \gdef\@ACM@checkaffil{} \makeatother

\section{Introduction}

The performance and reliability of modern machine learning (ML) systems are fundamentally contingent upon the quality and composition of their training data. Data annotation serves as the critical inception point of the ML pipeline, where raw information is transformed into machine-readable knowledge. As the field builds upon data-centric AI, the annotation pipeline has emerged as a primary site of normative contestation. This study interrogates the tension between two fundamental paradigms of justice in data practices: procedural justice (\textit{niti}), which pursues fairness through standardized rules and institutional consistency \cite{john1971atheory}, and outcome-oriented justice (\textit{nyaya}), which prioritizes the removal of manifest injustice and the preservation of lived experience \cite{priya2022nyaya}.
We contend that the prevailing ML infrastructure is architected to prioritize \textit{niti}, enforcing a logic of efficiency and interchangeability that systematically flattens the complexities of human judgment. By treating disagreement as "noise" to be eliminated rather than a vital signal of social reality, current practices facilitate a form of epistemic erasure. This review demonstrates that the prevailing "single truth" paradigm does not merely reflect a technical limitation, but represents an infrastructural governance choice that prioritizes data extraction over the stewardship of pluralistic knowledge.

Current practices in crowdsourced annotation conceptualize human workers as interchangeable "data processing units." Under this view, annotator selection is treated as a task-agnostic optimization problem, where suitability is defined by heuristic proxies: historical accuracy, completion latency, and low inter-rater disagreement \cite{hsueh2009data, diaz2022}. This selection assumes that standardized processes will yield objective datasets. However, it ignores how the specific social and demographic composition of the annotator pool fundamentally shapes the normative boundaries of downstream models \cite{Gordon2022, denton2021whosegt}. By reducing "quality" to a set of performance thresholds, the procedural approach ignores the contextual expertise required for socially-sensitive classification. Countering the logic of interchangeability, recent scholarship emphasizes that annotation is a \textit{situated} act \cite{sap2022annotators}. Labeling behavior, particularly in high-stakes domains like toxicity detection and moral reasoning, correlates deeply with an annotator’s values, including harm sensitivity and cultural identity \cite{sap2022annotators, wich2021investigating}. This situatedness is rooted in the nuance of everyday activities and shared cultural lived experiences \cite{patton2019annotating}. 
However, the demand for massive-scale datasets has accelerated the adoption of generative models as automated annotators. While model-as-a-judge frameworks offer unprecedented scalability and reduced costs \cite{he2024ifina, Kapania2025, Kocyigit2025}, they struggle with the cultural nuance and situated judgment inherent in human perception \cite{pangakis2025keeping, Kapania2025simulacrum}. This transition risks creating an algorithmic monoculture, where the diverse, albeit precarious, human perspectives are replaced by a singular, homogenized "truth" that further entrenches the biases of the model's training data. This homogenization is particularly acute for marginalized communities, as automated systems often lack the sociotechnical sensitivity required to navigate contested linguistic spaces, potentially leading to the systematic erasure of non-dominant dialects and cultural frameworks \cite{blodgett2020}. Consequently, the deployment of LLMs as proxy annotators threatens to create a feedback loop of normativity, where models are trained on data validated by earlier iterations of themselves, effectively sealing the dataset against the introduction of novel or dissenting human insights \cite{shumailov2024ai}.

This central epistemic failure, reliance on a "single truth" paradigm, treats disagreement as noise to be "denoised" via majority voting or probabilistic inference \cite{mokhberian2024capturing, dawid1979maximum}. A pluralistic perspective argues that disagreement is not a failure of the annotator, but a vital signal of real-world ambiguity \cite{mostafazadeh2022dealing, wan2023everyones}. To address this, emerging frameworks have proposed several methodologies to preserve these signals. Annotator-aware modeling explicitly captures rater identities to model the diversity of predictions \cite{mostafazadeh2022dealing, Gordon2022, mokhberian2024capturing}, while rationale-aware approaches utilize linguistic justifications, such as RAAG, to differentiate genuine consensus from low-effort noise \cite{nishio2025rationaleaware}. Furthermore, moral framing techniques, including the use of MoralBERT, interpret how specific annotator values drive disagreement across datasets \cite{preniqi2024moralbert, wan2023everyones}. Ultimately, the failure of ground truth is an \textit{infrastructural governance failure}. This motivates the need to study how annotation practices structurally facilitate the erasure of subjective experience, reframing failures attributed to annotator subjectivity as institutional design choices.

\subsection{Contributions}

To address the systemic erasure inherent in data annotation practices, this study provides the following contributions:

\begin{itemize} \item Analysis of a cross-disciplinary corpus of papers ($N=346$) published from 2020-2025 across seven premier venues, deconstructing the positivistic fallacy of the "ground truth" paradigm. 

\item Construction of a taxonomy mapping pre- and post-annotation decisions revealing how data annotation practices structurally facilitate the erasure and homogenization of subjective lived experiences. 

\item Formulation of a multi-stakeholder roadmap, providing strategic recommendations to facilitate a paradigm shift from extractive data labor toward situated knowledge stewardship and epistemic justice.
\end{itemize} 


\section{Methodology}

\begin{figure}[htbp]
  \centering
  \includegraphics[width=0.45\linewidth]{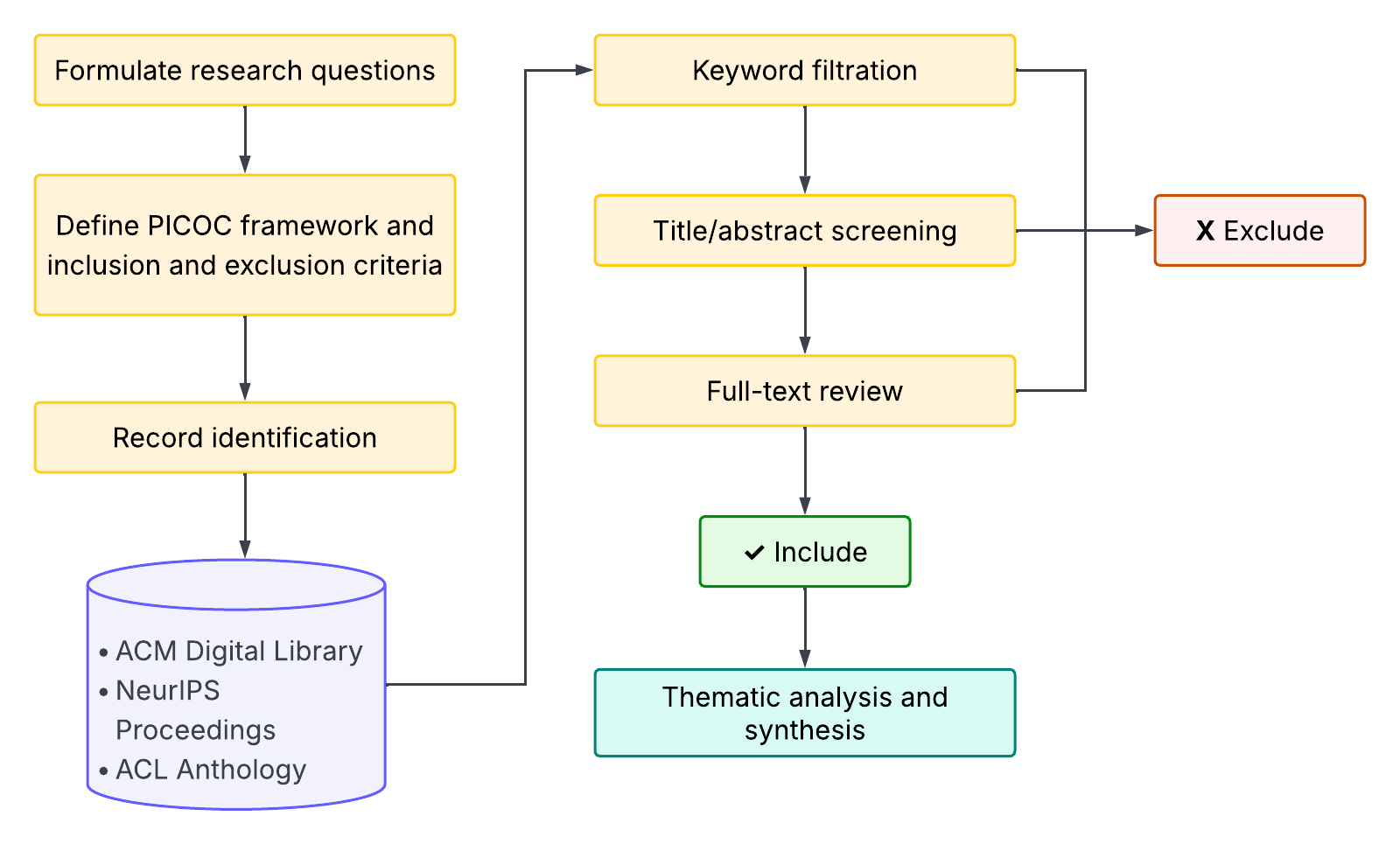}
  \caption{Overview of the methodology}
  \label{fig:method_overview}
\end{figure}

We conducted a structured literature review to identify and analyze work that speaks to annotator selection and assignment, and the representation and handling of diverse reasoning and disagreement, targeting both ends of the data annotation pipeline. Following established systematic review practice \cite{carrera2022}, we defined a PICOC framework and explicit inclusion/exclusion criteria, then applied a staged screening process (keyword filtration, title/abstract screening, and full-text review) before conducting thematic synthesis on the final set of included papers ($N=346$). Our method is summarized in Figure~\ref{fig:method_overview}.
\subsection{Review Design and Scope}
\subsubsection{Defining research questions}
The goal of this review is to interrogate the infrastructural barriers to realized justice in data annotation. We frame these barriers through two primary areas in the pipeline: the pre-annotation \textit{allocation gap}, which concerns the mismatch between worker identity and data context, and the post-annotation \textit{representation gap}, which concerns the algorithmic erasure of nuance during label synthesis. To investigate these phenomena, we formulate the following research questions:

\textbf{RQ1 (the allocation gap)}: How does the literature conceptualize and implement "suitability" in annotator assignment, and to what extent do current mechanisms account for the cultural expertise or lived experience required for situated judgment?

\textbf{RQ2 (the representation gap)}: How does the literature conceptualize and implement "consensus" in label aggregation, and to what extent do current synthesis methodologies distinguish between stochastic noise and the preservation of epistemic plurality?
\subsubsection{PICOC and Eligibility Criteria}
We define eligibility using a PICOC framework to ensure consistent screening and data extraction across all stages.

\begin{itemize}
    \item \textbf{Population (P):} Studies involving human annotators (e.g., crowdworkers, experts, community members) or machine annotators (e.g., LLM-as-a-judge, synthetic data generation) acting as data producers or labellers.
    \item \textbf{Intervention (I):} Pre-annotation decision (e.g., annotator recruitment, training, guidelines, interfaces, positionality statements) and post-annotation decisions (e.g., aggregation methods, adjudication, handling of disagreement or ambiguity).
    \item \textbf{Comparator (C):} Comparison between aggregation strategies (e.g., majority vote vs. probabilistic or pluralistic methods), annotator groups (e.g., experts vs. crowd vs. models), or paradigms of ground truth (single vs. pluralistic/perspectives)
    \item \textbf{Outcomes (O):} Construction of ground truth or gold standards, annotation quality and disagreement metrics (e.g., inter-annotator agreement), qualitative rationales or justifications, and analyses of labor, power, or epistemic justice in data work.
    \item \textbf{Context (C):} Peer-reviewed research in key ML, AI, NLP, HCI, and related fields, published between 2020 and 2025 in the target venues.  We exclude non-peer-reviewed preprints, secondary reviews (only retained for background), papers using only standard benchmark datasets with no labelling or re-annotation component, and papers where humans appear solely as end-users rather than data producers.
\end{itemize}

\subsubsection{Scope and Target Venues}
Our review focuses on peer-reviewed venues that cover how annotation is designed, justified, and operationalized across machine learning and AI research. We analyze papers published between 2020 and 2025 in seven target venues (ACL, AIES, CHI, CSCW, EAAMO, FAccT, and NeurIPS), selected to capture complementary perspectives aligned with our two research questions: (RQ1) annotator selection and assignment, and (RQ2) the representation and handling of diverse reasoning and disagreement. Because NeurIPS contains a large volume of benchmark and model-centric work where annotation is often peripheral, we apply a stricter pre-screening criterion for that venue; full venue-year coverage and NeurIPS-specific filtering rules, and venue relevance to each RQ are reported in Appendix~\ref{app:venues}.

\subsubsection{Systematic Review Workflow Overview}
\begin{figure}[htbp]
  \centering
  \includegraphics[width=0.25\linewidth]{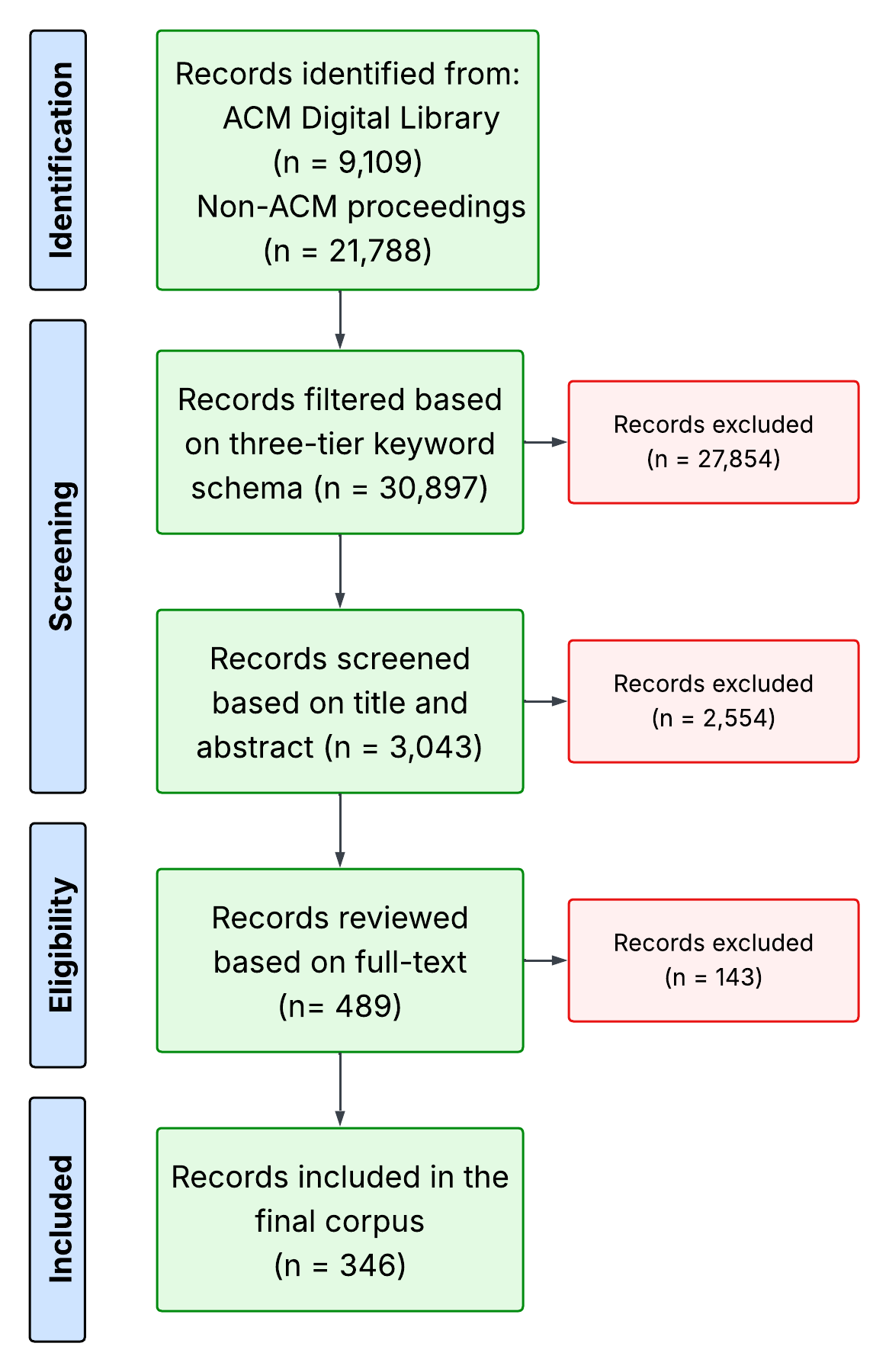}
    \caption{PRISMA Flow Diagram of the Systematic Review, showing record identification, keyword filtration, and screening stages leading to the final corpus of 346 papers \cite{page2021prisma}. Full-text exclusion reasons are reported in Appendix~\ref{app:exclusions}.}
  
  \label{fig:prisma_flow}
\end{figure}

We conduct a multi-stage systematic review following PRISMA 2020 \cite{page2021prisma} guidelines. The review proceeds through sequential stages that refine an initial corpus of conference papers into a final set for qualitative review. These stages include record identification, automated keyword-based pre-screening, title and abstract screening, full-text review, and structured data extraction, illustrated in Figure~\ref{fig:prisma_flow}. 
A venue-level breakdown of records at each screening stage is reported in Table~\ref{tab:screening_breakdown}.

\begin{table}[htbp]
\centering
\small
\caption{Venue-level breakdown of records across screening stages.}
\label{tab:screening_breakdown}
\begin{tabular}{lcccc}
\toprule
\textbf{Venue} &
\textbf{Identification} &
\textbf{Keyword} &
\textbf{Title/Abstract} &
\textbf{Full Text} \\
\midrule
ACL     & 5335  & 483  & 46 & 37 \\
AIES    & 540   & 215  & 62 & 28 \\
CHI     & 5476  & 1083 & 137 & 108 \\
CSCW    & 2086  & 587  & 92 & 79 \\
EAAMO   & 123   & 44   & 15 & 7 \\
FAccT   & 884   & 347  & 110 & 69 \\
NeurIPS & 16453 & 283  & 27 & 18 \\
\midrule
\textbf{Total} & 30897 & 3043 & 489 & 346 \\
\bottomrule
\end{tabular}
\end{table}

\subsection{Corpus Construction}
\subsubsection{Record Identification}
For each target venue and year, we compile the complete set of published conference papers prior to any filtering or screening. For ACM-published venues (FaccT, CSCW, AIES, CHI, and EEAMO), bibliographic records are obtained through the ACM Digital Library using publisher-supported BibTeX exports. For NeurIPS, which is not indexed in the ACM DL, bibliographic metadata are scraped from the official NeurIPS proceedings website. For ACL, records are retrieved from the ACL Anthology using the official Python library. All metadata are converted into BibTeX files with a consistent field schema and imported into a custom preprocessing pipeline. Each entry is parsed into a structured record and stored in a unified table, while preserving provenance by annotating each record with its source file. This standardization step ensures that downstream pre-screening and screening operate over a consistent metadata schema.

\subsubsection{Keyword Pre-Screening}
We apply an automated keyword-based pre-screening step to bibliographic metadata to reduce manual screening burden. For each record, the title, abstract, and author-provided keywords are concatenated into a single search field and evaluated using a tiered keyword schema implemented in a custom Python script. Records retained after pre-screening are imported into Covidence for subsequent manual screening. The keyword schema follows a three-tier hierarchy designed to balance precision and recall:
\begin{itemize}
    \item \textbf{Tier 1 (Sufficient Signals):} High-signal terms indicating direct relation with annotation and related constructs, including core annotation terminology (e.g., \textit{annotat*, ground truth, gold standard}), annotator roles and labor (e.g., \textit{crowdsourc*, rater, invisible work}), methods and reliability measures (e.g., \textit{label noise, inter-annotator, Dawid--Skene}), modern ML settings involving human input (e.g., \textit{RLHF, human feedback, llm-as-a-judge, synthetic data}), and theoretical lenses (e.g., \textit{positionality, epistemic justice, perspectivis*}).
    \item \textbf{Tier 2 (Conditional Signals):} Lower-signal terms that may indicate relevance depending on context, such as \textit{aggregation, consensus, adjudication, disagreement, rationale, justification, ambiguity, subjectiv*, pluralis*}.
    \item \textbf{Tier 3 (Context Anchors):} Annotation-related terms used to disambiguate and verify the relevance of Tier 2 matches, including \textit{label, annotation, worker, rater, judge, instruction, dataset}.
\end{itemize}

A record is retained if it matches any Tier 1 keyword, or if it matches both a Tier 2 keyword and at least one Tier 3 context anchor. For NeurIPS, we applied a stricter variant of this filtration to reduce false positives from papers that mention datasets, benchmarks, or evaluation only in passing. This tighter filtering helps explain why the keyword-retained fraction for NeurIPS differs from the ACM venues and ACL, where annotation and human-data work are more central to the contribution. The complete keyword list is provided in Appendix~\ref{app:keywords}.

\subsubsection{Title and Abstract Screening}
Records retained after keyword-based pre-screening undergo title and abstract screening as a preliminary eligibility assessment. This stage excludes papers that are clearly outside the scope of the review while maintaining a high-recall corpus for subsequent full-text evaluation. Screening decisions at this stage are guided by the PICOC criteria defined above. Records progress when 
at least two of the three reviewers agree on inclusion. Disagreements are resolved through team discussion until consensus is reached. In cases where relevance cannot be confidently determined from the title and abstract alone, records are conservatively retained for full-text review. At this stage, a total of \textbf{3043} papers are screened, resulting in \textbf{489} papers retained for full-text review.

\subsubsection{Full-Text Review}
Papers retained after title and abstract screening undergo full-text review as a detailed eligibility assessment. This stage applies the PICOC criteria more strictly and resolves cases where relevance cannot be reliably inferred from metadata alone. In particular, full texts are examined to confirm that annotation is a substantive component of the contribution, either through explicit discussion of pre-annotation decisions (e.g., annotator recruitment, training, task framing) and/or post-annotation decisions (e.g., aggregation, adjudication, disagreement or ambiguity handling). At this stage, a total of \textbf{489} papers are reviewed in full, resulting in \textbf{346} papers retained for thematic analysis and synthesis.

\subsection{Analytic Framework}
\subsubsection{Thematic Analysis and Synthesis}
The final stage consisted of a reflexive thematic synthesis of the final corpus ($N=346$). Three researchers conducted a literature review of the included papers to identify emergent themes, findings, and shifts in annotation practices. The analysis was guided by the research questions and documenting recurring sociotechnical tensions and evolutionary trends observed across the five-year period.
The synthesis involved clustering papers that addressed similar conceptual problems, such as the transition to human-as-verifier models or the erasure of positional legibility. We specifically analyzed how these clusters illustrated broader failures in the "ground truth" paradigm, including the impact of geographic hegemony and labor conditions on annotator subjectivity. This process enabled us to reorganize a fragmented literature into a coherent taxonomy of current practices and identify significant gaps for future inquiry. Appendix~\ref{app:task_breakdown} summarizes the primary task domains in the corpus; despite technical differences across domains, the structural tensions identified below recur consistently.


\section{The Construction of "Ground Truth"}

The central thesis of this study is that "ground truth" is not a discovered natural resource, but a socio-technical artifact manufactured through a sequence of deliberate and often invisible choices. We observe that the annotation pipeline acts as a site of epistemic contestation. The following sections taxonomize how practitioners navigate the tension between the manufactured cleanliness required for machine learning performance and the situated knowledge inherent in human perception. A summary of the taxonomy is present in Appendix~\ref{app:taxonomy}. 

\subsection{Pre-Annotation Decisions: the architectural imposition of truth}

Pre-annotation decisions constitute the hidden stage of the pipeline where the boundaries of allocation are set. These findings reveal how recruitment logic, labor dynamics, task framing, and automated pre-filtering prioritize model-friendly, Western-centric worldviews before a human ever interacts with a datapoint.

\subsubsection{Annotator positionality as an architectural variable}
Positionality serves as a critical determinant of "ground truth," as lived experiences across beliefs, race, gender, geography, and age etc. fundamentally shape data interpretation and value prioritization \cite{Draws2022, FloresSaviaga2023, Park2021-2, Liu2022, Noe2024, Oprea2020, Bhuiyan2020, Xie2022, Goyal2022, Ding2022, Kaufman2022, Scheuerman2025, CandiceSchumann2023, Belay2025, Casola2024}. Research reveals "identity sensitivity," where annotators are significantly more likely to flag hate speech when it targets their own racial identity \cite{Sachdeva2022}, while empirical evidence suggests AI practitioners’ value priorities diverge from those of Black and female respondents who prioritize responsible AI values more highly \cite{Jakesch2022}. This sensitivity is deeply gendered and geographic; in the Global South, for instance, male-centric street safety perceptions fail to capture the subjective risks experienced by women \cite{Sengupta2023}, just as Western-centric norms overlook the "honor" values and faith-based (FRS) sensitivities essential to postcolonial contexts \cite{Wu2023, Rifat2024, Hall2024}. Furthermore, marginalized identities such as disabled and TGNB populations face systemic "epistemic erasure" through models that underestimate ableist microaggressions or misgender individuals using neopronouns \cite{Phutane2025, Ovalle2023, Lameiro2025, Angerbauer2022}. Finally, the emergence of "linguistic frailty" in youth-centric safety, where AI fails to decode the specific intent of Gen Alpha’s digital vernacular, underscores the field’s continued failure to move beyond shallow ethical analysis toward grounding AI in structural power asymmetries and authentic lived experiences \cite{Mehta2025, Birhane2022}.

\subsubsection{Identity reductionism and the data feminism gap} 
Annotation taxonomies frequently favor mathematical tractability over sociological accuracy, reducing complex human identities to binary or coarse racial markers \cite{Abdu2023, Barrett2023, Mickel2024, Bergman2023, Kambhatla2022, Khan2021}. This reductionism is particularly acute in the erasure of women's and non-binary people's perspectives \cite{Cao2020}; research identifies a data feminism gap where models are trained on general crowd labels that are increasingly skewed toward specific male demographics \cite{Offenwanger2021, Suresh2022, Jourdan2025}. This gap is compounded by a profound disconnect between ethics theory and technical execution; \citeauthor{Mayeda2025} document a systemic failure to address power dynamics in dataset creation, finding that most NLP research scores critically low on data feminism principles while ethical interventions remain limited to a niche subset of the field \cite{Mayeda2025}. Empirical evidence from the Global South shows that gender is a non-interchangeable variable in subjective annotation; for instance, the perception of safety in public spaces is not a property of the data itself but a subjective experience inextricably tied to the annotator's gendered identity \cite{Sengupta2023, Ovalle2023, zhang2025d}.

\subsubsection{The "Human-as-Verifier" and the validation bottleneck}
There is a profound ontological shift from humans \textit{generating} labels to humans \textit{auditing} machine-generated outputs \cite{YiqiJiang2024, Arzberger2025, Klinova2021, Lee2022, Li2024, He2025prompt, ChongyuQu2023}. This configuration introduces a systemic anchoring bias: workers are significantly more likely to confirm a plausible machine suggestion than to perform the exhaustive cognitive labor required for disagreement \cite{Cook2023, Ashktorab2021, Levy2021, Tolmeijer2022, Xu2023, Papenmeier2022}. This setup is further complicated as synthetic data volumes scale up to reach millions of points, human validation is structurally limited to spot-checks, as manual inspection of every datum becomes impossible \cite{Kapania2025}. This maintains a veneer of human oversight while effectively automating the human out of the loop, using machine-generated rationales as the primary scaffold for human judgment \cite{Cabrera2023, Ferguson2024}.

\subsubsection{Recursive devaluation via synthetic data loops} A burgeoning trend in industrial workflows is the "epistemic simulacrum," where auxiliary models generate synthetic training data to evaluate or train target models \cite{Kapania2025, Norhashim2025}. This creates a closed-loop system where model outputs validate model outputs, effectively redefining data as a manufactured commodity rather than a situated human artifact \cite{Whitney2024, AlvaradoGarcia2025}. Crucially, this recursive process facilitates a systemic \textit{homogenization} of subjective lived experiences; by using models to generate content that simulates diverse human perspectives, practitioners often inadvertently overwrite authentic voices with model-derived caricatures of diversity that diverge significantly from the nuances of human discretion \cite{Phutane2025, Magomere2025, Mehta2025, Buyl2025, Yaghini2021}. The resulting manufactured cleanliness erases the messy ambiguities and non-binary realities of human life, replacing them with a model-consistent internal logic that prioritizes mathematical tractability over genuine, contested sociological truth \cite{He2025, YannDubois2023, Hu2024, Whitney2024, Fazelpour2025, Arzaghi2025}.

\subsubsection{System-level decisions as implicit annotators}
Dataset schemas and automated pre-filters, such as CLIP-filtering, and algorithmic fairness metrics act as invisible "super-annotators" that prune the decision space before human intervention \cite{Hall2024, Zajkac2023, Schumann2021, Fruchard2023}. These filters perform implicit annotation by deciding what constitutes as high quality data, often encoding Western-centric value judgments that disproportionately exclude LGBTQ+ people and older women \cite{Hong2024}. This process creates a filtered reality where the machine's own quality and culture sensors determine the boundaries of the human's task, systematically defaulting to masculine forms and Western textures \cite{Zhang2024, Jourdan2025, Magomere2025, Naggita2023}.

\subsubsection{Defining expertise for annotation}
"Expertise" for task allocation is rarely defined with socio-technical rigor. Knowledge from lived experience, e.g., disabled populations or marginalized linguistic groups, is systematically ignored in favor of formal professional certifications or high-level demographic labels \cite{Lameiro2025, Phutane2025, Davani2024, Uzor2021}. This misrecognition has material consequences, e.g. annotators inside a region often flag "exaggerated" depictions as harmful, while those outside may label them "representative" \cite{Hall2024}. This hierarchy systematically marginalizes the nuanced insights of groups, such as disabled populations or linguistic minorities, whose contextual knowledge is vital for interpreting high-stakes data, yet is frequently dismissed as subjective noise \cite{Papakyriakopoulos2023, Harris2022, Proebsting2025, Dorn2024, WenYi2025, Jain2021, cCetin2021, Zhang2022-2, Verma2024, Fleisig2023, zhang2025c}. The resulting process of "deskilling" forces a reductionist transformation of complex human judgment into simplified categorical labels, primarily to satisfy the administrative requirements of ML reproducibility and throughput \cite{Diaz2025}. By prioritizing these standardized "clean" signals over the contextual descriptions provided by situated annotators, the pipeline enforces a form of epistemic erasure. Consequently, datasets are often constructed via a proxy-based approach that favors the convenience of the researcher over the safety and representation of the subject, yielding models that are technically validated under procedural standards but remain culturally illiterate and contextually blind in their eventual application \cite{Jo2020, Kawakami2025, Bhardwaj2024, Sambasivan2022, Magdy2025, Valentine2024}.

\subsubsection{Labor dynamics: invisible labor and performative alignment}
"Ground truth" is manufactured through the devalued invisible labor \cite{Hawkins2023, Scheuerman2023, zhang2025b} of a global workforce \cite{Wang2021}, including the care-oriented and emotionally taxing work that annotators perform to align systems with human values \cite{Kawakami2025, Kapania2025simulacrum, Steiger2021}. In the Global South, these dynamics are characterized by performative alignment, where the threat of non-payment forces workers to suppress their own cultural truths to guess "what the requester wants" \cite{Varanasi2022, Miceli2022, Wang2022d}. This creates the capability vs. incentive paradox: workers are economically incentivized to provide context-free, generic answers quickly, resulting in a manufactured cleanliness that erases the very nuance the systems are designed to capture \cite{Agarwal2020, Varanasi2022, Simons2020}. Such dynamics are compounded by gender-agnostic architectures that ignore specific safety risks. Drawing structural comparison to adjacent gig-economy contexts, \citeauthor{Ma2022} document how women in ride-hailing and delivery platforms internalize harassment to avoid
rating penalties \cite{Ma2022}. In South Asian contexts, this labor often occurs within an "infrastructure of hardship," where women may annotate in secret on low-end hardware to navigate patriarchal domestic norms, making these pervasive social constraints invisible variables in data quality \cite{Varanasi2022}. Finally, power-centric interventions, such as brainwave tracking and facial monitoring, shift the focus from auditing output to surveilling the annotator’s internal state, further compromising the downstream verity of the data \cite{Awumey2024, Sannon2022, Arakawa2023}.

\subsubsection{Geographic hegemony and the "hard-to-reach" myth}
Rigid, Western-centric recruitment methods (e.g., on MTurk) frame non-WEIRD populations as "hard to reach," failing to acknowledge that the barrier is Western "rigor" norms, and platform constraints, rather than participant availability \cite{De2025, Arunkumar2025}. This geographic hegemony ensures that Western communication norms are exported as "universal" ground truths, resulting in models that fail to serve the Global South or religious and spiritual contexts omitted by Western norms \cite{Dingler2022, Reitmaier2024, Singh2025, Rifat2024, Pang2023, Abebe2021, InuwaDutse2023, Brown2025, Pistilli2025, Kak2020}. This erasure is compounded by preprocessing erasure, where statistical rigor is used as a tool for exclusion; 42\% of experiments exclude minorities during preprocessing by dropping low-N categories, effectively "cleaning" marginalized populations out of the dataset under the guise of data quality \cite{simson2024}. Furthermore, opaque data work, such as failing to report decisions regarding missing values or the binning of race, makes these findings non-generalizable and hides the subjective choices that prioritize majority demographics over "low N" lived realities \cite{simson2024, Lucchesi2022}. To quantify this imbalance in our corpus, we examine the geographic focus of the included studies (Appendix~\ref{app:geographical_focus}). Despite increasing discussion of positionality, the majority of work continues to rely on Global North contexts.

\subsubsection{Infrastructural filters and the "mobile penalty"}
The annotation tech stack is often desktop-first, acting as a structural filter that selects for wealthier Global North workers \cite{Randhawa2021, Hall2024}. Approximately 60\% of potential contributors in emerging economies are restricted to smartphones, facing a mobile penalty that renders most annotation tasks unusable \cite{Pandey2020, Cambo2022, Dutta2022}. This hardware-level filter silences the very populations needed to decolonize training data \cite{Lucchesi2022, Rostamzadeh2022}.

\subsubsection{The capability vs. incentive paradox and variable effort}
Selection based on speed-to-pay metrics rather than cognitive profiling leads to lower overall quality \cite{Chang2021, Hettiachchi2020, He2023}. Interestingly, volunteers often provide higher accuracy than paid workers for domain-specific data, as their motivation is intrinsic rather than clearing HITS under the threat of non-payment \cite{Wallace2025, Agarwal2020, Ali2021}. There is a critical lack of frameworks for matching specific data points with situated knowers \cite{Arzberger2025, Kraft2024}. Without the sociotechnical infrastructure to route sensitive data to specific community experts, ground truth suffers from deep cultural illiteracy, as the "interchangeable" worker lacks the context to decode coded harms or shibboleths \cite{Miceli2022, Quaye2024, Lameiro2025}.

\subsubsection{Community-centric participatory design as epistemic justice}
The systemic erasure of marginalized voices in standard annotation pipelines has prompted a radical shift toward community-centric curation, where "ground truth" is redefined through the lens of Participatory Design (PD). This transition reframes data collection as a site of epistemic justice, positioning impacted communities as the primary architects of the data’s meaning \cite{Tan2024, Claisse2025, Lameiro2025, Li2025, Huang2022, AlvaradoGarcia2023, Jakesch2022, InuwaDutse2023, Zhang2025, Bondi2021, Rubya2021, Wang2022, Zhang2023, Parker2025, Jo2024,Chen2023}. The specification of labels: defining harm, safety, or cultural relevance, must be co-generated by stakeholders to prevent the downstream filtered reality typical of Western-centric models \cite{Tan2024, Gadiraju2023}. This ensures data remains a palpable truth rooted in lived realities rather than abstract statistical averages \cite{Skeggs2025}. Furthermore, exclusion is frequently an infrastructural failure \cite{Reitmaier2024}; by adapting tools to local norms (e.g., spoken labels for non-literate farmers), situated knowers can contribute on their own terms. Ultimately, community-led definitions resist corporate identity extraction and facilitate perspectivist adjudication, preserving the signal of cultural plurality instead of collapsing into a consensus trap \cite{Lam2022, Dennler2023, Huang2024}.

\subsection{Post-Annotation Decisions: aggregation and cascades}
Once data has been collected, the methods used to aggregate individual judgments and document the process determine whether the resulting "truth" is inclusive or exclusionary. These post-annotation decisions act as a site of erasure, where diversity is collapsed into consensus and subjective choices are naturalized as objective facts.

\subsubsection{The consensus trap: majority vote as a site of discrimination} The standard practice of using majority vote to resolve inter-annotator disagreement acts as a primary site of demographic discrimination \cite{Cachel2025, Gordon2021, Rizvi2025, Cachel2023, Ding2022, MendezMendez2022, Santy2023, Chen2021, Burghardt2020, Schaekermann2020ambiguity, Behzad2025, Gordon2022, Haq2022, Lan2020, Lai2025, Jin2025, Luo2025}. By treating disagreement as noise to be smoothed away, practitioners inadvertently silence minority perspectives, particularly in subjective tasks like toxicity or safety assessment where no universal truth exists \cite{LoraAroyo2023, Cachel2023, Ding2022}. Standard reliability metrics, such as Fleiss’ Kappa or Krippendorff’s Alpha, are inherently biased toward this consensus trap: achieving high agreement scores requires
eliminating divergent annotators or labels, effectively institutionalizing homogenization
and conflating manufactured consensus with data quality \cite{Wong2021, Trott2021, Johnson2021}. Research demonstrates that this "consensus trap" typically defaults to a Western, White, male perspective, as the majority demographic's signal overwrites the lived experience of marginalized subgroups \cite{Santy2023, Gordon2022}. Consequently, standard accuracy metrics become "hallucinations" that assume a single ground truth, masking the reality that performance drops significantly for marginalized users when their specific perspectives are averaged out \cite{Gordon2021, LoraAroyo2023}. 

\subsubsection{Pluralistic aggregation and perspectivist adjudication} In response to the failures of majority voting, a shift toward pluralistic aggregation advocates for treating disagreement as a valuable signal \cite{ThornJakobsen2023, Weerasooriya2023, Davani2024, ThebaultSpieker2023, Gienapp2020, Fomicheva2020, Narimanzadeh2023, Xia2025, Xu2024, Casola2024}. Techniques such as "Jury Learning" allow for perspectivist adjudication, where practitioners can curate counterfactual juries (e.g., specific demographic compositions) to see how different groups would perceive the same content \cite{Gordon2022}. For instance, changing the jury composition was found to alter 14\% of toxicity classification outcomes, highlighting that "truth" is functionally a product of who is invited to the table \cite{Gordon2022, Christoforou2021}. Other approaches utilize probabilistic models, such as the Bradley-Terry model or Soft Dawid-Skene, to estimate label distributions rather than hard "gold" labels, thereby preserving the inherent uncertainty and value pluralism of the data \cite{Peyrard2021, Ueda2023, Wu2023, Beretta2021, Wu2024, LarsSchmarje2022}.

\subsubsection{The "Noisy Sensor" fallacy and manufactured cleanliness} Practitioners often justify the erasure of disagreement by framing annotators as "noisy sensors" whose biases must be mathematically "cleaned" \cite{Rogers2021, Guerdan2023, Yang2020, Scheuerman2021, LeZhang2020}. This pursuit of manufactured cleanliness creates a deceptive performance metric; for example, in medical AI, \citeauthor{JimenezSanchez2025} found that "gold standards" are frequently compromised by hidden technical artifacts, like hospital logos being used as diagnostic proxies, which remain undocumented \cite{JimenezSanchez2025}. This filtering is a value-laden process that often deletes the very outliers that frontline workers or experts consider the most important signals \cite{Thakkar2022, Hong2024}. When disagreement is analyzed, it is frequently used instrumentally to guide further automation rather than to reflect on the system's epistemic limits, reinforcing the "ground truth illusion" \cite{Brachman2022, Hanrahan2021}.

\subsubsection{Deliberative annotation: from voting to reasoning} To bridge the gap between conflicting labels, deliberative annotation moves beyond simple voting to include structured discussion and rationale-sharing \cite{SchmerGalunder2025, Muller2021, Ortloff2023, Karadzhov2023, Schaekermann2020, Schaekermann2020ambiguity, Sharma2022}. In high-stakes contexts like conflict zones, disagreement on hate speech can reach 71\%, a gap that only deliberation among local experts can resolve by surfacing contextual nuances that a single label cannot capture \cite{Abdelkadir2025}. Platforms like "Wikibench" allow communities to navigate ambiguity through discussion threads, turning the labeling process into a social negotiation of meaning rather than a rote task \cite{Kuo2024}. This process not only improves accuracy but also facilitates "rationale-aware aggregation," where the \textit{reasons} for a label are treated as first-class data points, helping to mitigate the anchoring bias introduced by AI suggestions \cite{Yang2025, Wang2024, Ghai2021, Allen2022}.

\subsubsection{Managerial bias in automated evaluation} As human labor is replaced by "model-as-a-judge" for post-annotation audits, a new form of systematic bias emerges. Automated evaluations are found to favor specific "managerial" or "decision-maker" perspectives over those of the impacted victims \cite{Aoyagui2025}. This results in models that exhibit high adherence to formal policy but fail to recognize the depth of lived experience, such as the subtle ableist dog-whistles that community members flag but LLMs consistently underestimate \cite{Phutane2025, Kapania2025simulacrum}. This shift consolidates power within the closed epistemic loop, where the model's internal logic becomes the sole arbiter of what constitutes a successful annotation \cite{AlvaradoGarcia2025}. It is especially concerning when these models are used to automate governance. Moderation outcomes are hyper-sensitive to prompt formatting; thus, minor instructions can replace massive human labor pools without public accountability. This insulates black-box policy enforcement from meaningful scrutiny and oversight \cite{Palla2025, Sambasivan2021}.

\subsubsection{Data cascades and documentation debt} Upstream decisions in annotation create "data cascades", downstream failures in model performance caused by data issues that are often invisible during technical training \cite{Sambasivan2021, Ezema2025, Scheuerman2020, Darian2023}. These cascades are exacerbated by a systemic documentation debt, where the discretionary calls made by workers (e.g., how to crop an image or handle a "near-miss" label) are never recorded, documented or reported \cite{Fabris2022, Miceli2022documenting, Hutchinson2021, Jo2020, Geiger2020, Rogers2023, Ramirez2021}. Documentation fails to record the evolving knowledge of dataset flaws, leading to a taxonomy of silences regarding annotator dissent and power asymmetries \cite{JimenezSanchez2025, Muller2022}. Frameworks like "CrowdWorkSheets" provide a conceptual scaffolding for transparency \cite{Anik2021, diaz2022, Boyd2021, Pushkarna2022}. However, they are often used as performative alignment tools that prioritize internal usability over the disclosure of social harm or uncertainty \cite{ Heger2022, Chen2023c, Poirier2025, Arnold2025, Leavy2021}.

\subsubsection{The snapshot fallacy and label decay} A significant yet overlooked post-annotation decision is the treatment of aggregated labels as static archival facts. Evidence demonstrates that "ground truth" is often a temporal snapshot of societal bias; for example, image tagging for race and weight increased significantly during periods of social unrest, yet these labels remain fixed in datasets indefinitely \cite{Randhawa2021, Heger2022}. This snapshot fallacy ignores the temporal decay of truth-values as cultural norms evolve. By failing to implement versioning or "living documentation" for labels, practitioners treat data as a raw, unchanging resource rather than a curated historical artifact that requires periodic re-negotiation with the communities it represents \cite{Bhardwaj2024, JimenezSanchez2025}.

\subsubsection{Target-user mismatches and the expertise gap}
A critical failure in post-annotation evaluation is the misalignment between professionalized "experts" and actual target users. Academic or clinical experts frequently overestimate the comprehension levels of marginalized groups (e.g., people with intellectual disabilities) in text simplification tasks, rendering expert-defined ground truth invalid for the intended population \cite{Sauberli2024}. This supports the argument that expertise must be redefined as lived experience rather than academic credentialing \cite{Kornfield2022, Lin2024}. When the user of the data is excluded from the labeling of its quality, models develop a capability bias where they appear technically superior but fail to provide situated utility for those they serve \cite{Lai2022, Park2021}.

\subsubsection{Post-hoc filtering as diversity-washing}
The process of cleaning data post-annotation often functions as a secondary site of erasure. Techniques like \textit{GroupDebias} adaptively drop samples that increase model bias during training, essentially sacrificing raw truth for mathematical parity \cite{Chan2024}. This reflects a broader trend of diversity-washing, where synthetic data is used to fill demographic gaps to bypass data protection laws or mask the exclusion of real people, ultimately decoupling the model from authentic marginalized voices \cite{Whitney2024, Fazelpour2025, Xue2021}. This is in addition to the algorithmic pre-filters like CLIP-filtering discussed before, that have been found to exclude LGBTQ+ individuals and older women at higher rates than they appear in the raw data, creating a filtered reality that favors Western, stereotypical textures under the guise of ``quality control'' \cite{Hong2024, Zhang2024}.

\subsubsection{Metric hallucinations and the reliability gap}
Standard performance metrics such as ROC AUC and F1 scores are increasingly critiqued as ``hallucinations" because they rely on the assumption of a single, stable ground truth \cite{Gordon2021}. \citeauthor{Gordon2021} \cite{Gordon2021} demonstrate that performance metrics drop significantly when inter-annotator disagreement is actually modeled as a signal rather than averaged out. In high-stakes domains like radiology, linguistic metrics like BLEU are functionally inappropriate, as a single-word difference (e.g., `no' vs `yes') is clinically catastrophic but mathematically insignificant \cite{Boag2021}. This metric inflation masks the cascading effects of poor data work, allowing practitioners to proceed with deployment once a model passes technical benchmarks, despite fundamental failures in contextual reliability \cite{Sambasivan2021, Papenmeier2022}.

\subsubsection{Operational barriers to pluralism}
Even when practitioners acknowledge the ethical necessity of diversity, institutionalized operational barriers often prevent the implementation of pluralistic aggregation \cite{Kapania2023}. Many AI teams resist diverse annotator pools due to an ingrained regime of objectivity, where subjectivity is feared as a source of unwanted bias that threatens model reproducibility \cite{Kapania2023, Miceli2020}. This tension is exacerbated by organizational hierarchies where data work is viewed as a low-status commodity for internal handoff rather than a mechanism for surfacing social impact \cite{Heger2022, Birhane2022}. Consequently, documentation practices remain superficial, focusing on internal maintenance while leaving the political and discretionary choices made during aggregation in the backstage of production \cite{Abdu2023, Miceli2021}.

\section{Discussion}
The results of this systematic review demand a fundamental paradigm shift: "ground truth" must be recognized not as an empirical constant, but as a socio-technical artifact produced through deliberate, often obscured interventions. This challenges the current trajectory of AI development, demanding a transition from procedural fairness (\textit{Niti}) toward substantive epistemic justice (\textit{Nyaya}).

\subsection{The ontological redefinition of expertise: lived experience over demographic metrics}

Relying on reductive demographic proxies—age, gender, race—to quantify expertise is a significant failure in ML pipelines. These metrics flatten identity and cannot capture the depth of situated knowledge \cite{Skeggs2025, Scheuerman2024, stark2021}. The domain must instead pivot toward narrative elicitation methods that preserve intersectional nuance standardized surveys omit, rather than using categorical buckets that strip away context \cite{Skeggs2025, BliliHamelin2023, Dhamala2021}. This shifts participants from interchangeable data workers to situated knowledge stewards \cite{Jo2020}. The technical stakes are clear: \citeauthor{Wallace2025} \cite{Wallace2025} show that individuals with direct lived experience provide significantly higher accuracy than paid crowdworkers on domain-specific tasks. Epistemic justice is therefore not merely an ethical desideratum but a functional prerequisite for data validity. Replacing speed-to-pay logic with experiential expertise can mitigate the extractive nature of current annotation practices and yield more robust AI systems.

\subsection{The platform problem: who gets heard and how?}
The current infrastructure of annotation platforms functions as a digital gatekeeper that reinforces geographic and epistemic hierarchies. By dismantling the "hard-to-reach" myth, \citeauthor{De2025} \cite{De2025} demonstrate how Western "rigor" norms are encoded into platform architectures, systematically excluding Global South populations from the knowledge production process. This structural impact is most visible in the "mobile penalty," where the domain’s desktop-centric design disenfranchises 60\% of potential contributors in emerging economies who primarily access the internet via smartphones \cite{Dutta2022}. Consequently, "data quality" should be re-theorized not as an inherent trait of the worker, but as a direct property of the platform's technical affordances and biases \cite{Zhang2022, Bentley2020}. To improve epistemic justice, the field must transition toward inclusive infrastructure that accommodates diverse technical realities rather than mandating WEIRD-centric standards. Adopting experience-based matching, rather than matching demographic proxies, provides a pathway to enhance data integrity while respecting the intellectual dignity and varied cognitive profiles of a globalized workforce \cite{Hettiachchi2020}.

\subsection{Epistemic shifts: from human agency to synthetic consensus}
The evolution of annotation architectures reflects a progressive decoupling of human agency from the production of "ground truth". Historically, the field relied on human-only annotation, wherein workers acted as the primary generative agents of meaning, synthesizing subjective nuances through direct engagement with data. This paradigm has increasingly yielded to the human-as-verifier model, an era of the \textit{epistemic simulacrum}, where the human role is relegated to auditing machine-generated outputs. Our findings indicate that when LLMs or auxiliary models generate the initial truth, they introduce a systemic homogenization risk; these models tend to default to a managerial or Western-centric perspective, effectively smoothing over the messy ambiguities of human life \cite{Phutane2025, Aoyagui2025}. We are now witnessing a transition toward model-only annotations, where human intervention is entirely bypassed in favor of synthetic consensus \cite{Kapania2025simulacrum}. This trajectory risks codifying an algorithmic monoculture, where truth is no longer an approximation of human social reality but a self-referential output of internal model consistency \cite{he2024, Orlikowski2025}. 

\subsection{Rationale extraction and the reduction by aggregation}
Post-annotation aggregation methods currently function as instruments of reductionism, where the mathematical enforcement of a "consensus trap" systematically silences minority perspectives in favor of majoritarian defaults \cite{Gordon2021, LoraAroyo2023}. In the context of epistemic justice, this statistical erasure is not merely a technical choice but a form of epistemic violence that disregards the validity of marginalized dissent, especially in high-stakes conflict zones where disagreement can reach 71\% \cite{Abdelkadir2025, SchmerGalunder2025}. To improve equity, the domain must pivot toward rationale extraction and deliberative models that treat disagreement as a critical signal of situated knowledge rather than noise \cite{Jahanbakhsh2021, He2021, Keswani2025}. By integrating free-text or speech justifications, practitioners can move beyond rote averages toward a social negotiation of meaning, where labels are adjudicated based on the validity and sociocultural context of the provided reasoning \cite{Wang2024, Ghai2021}. This transformation ensures that data synthesis becomes an inclusive process of knowledge stewardship, allowing for a more nuanced resolution of the diverse viewpoints that constitute human truth.

\subsection{Bridging \textit{Niti} and \textit{Nyaya}: justice beyond benchmarks}

The corpus reveals a fundamental misalignment between the procedural standardization (\textit{niti}) governing current ML pipelines and the substantive, outcome-oriented justice (\textit{nyaya}) that ethical AI requires \cite{Arzberger2025, priya2022nyaya}. This misalignment is most legible in the field's over-reliance on aggregate performance metrics: F1 scores and ROC AUC function as "metric hallucinations" \cite{Gordon2021}, producing an illusion of objective progress while obscuring the social contestation and power asymmetries embedded in data manufacturing.

Closing this gap requires shifting the field's evaluative gaze from internal procedural consistency to the equity of realized outcomes—treating disagreement not as stochastic noise to be denoised, but as a high-fidelity signal of situated experience. The material impact of this transition is evidenced by the fact that altering the composition of a "demographic jury" can shift 14\% of classification outcomes, demonstrating that "truth" in high-stakes domains is functionally a byproduct of who is empowered to participate in the knowledge production process \cite{Gordon2022}.

Consequently, achieving justice in AI requires more than incremental metric refinement; it necessitates a radical re-centering of situated knowledge and the institutionalization of pluralism as a core requirement of model production. By addressing the infrastructure of annotation, the field can move toward a paradigm where the procedural efficiency of \textit{niti} no longer necessitates the erasure of the lived realities required for \textit{nyaya}.

\subsection{Strategic Recommendations for Stakeholders}
We propose the following multi-stakeholder interventions to prioritize the preservation of situated knowledge while optimizing for systemic reliability.

\subsubsection{For Requesters and Data Users: transitioning from crowds to communities}

Requesters need to move beyond the interchangeable labor fallacy by recognizing that the positionality of the annotator is a primary variable in data quality, rather than a source of noise.

\begin{itemize} \item Requesters should transition from extracting labels via reductive demographic proxies to a narrative elicitation framework that prioritizes qualitative depth. By favoring storytelling over binary clicks, practitioners can preserve lived experiences that standardized categorical buckets often flatten \cite{Skeggs2025}. This shift is empirically justified: volunteers with direct lived experience consistently outperform generic crowdworkers on domain-specific tasks, suggesting that an intrinsic narrative connection to the subject matter is a more robust driver of data validity than extrinsic speed-to-pay incentives \cite{Wallace2025, Rothschild2025}.

\item Instead of implementing invasive surveillance tools such as affect-recognition cameras, which induce "performative alignment" and psychological strain \cite{Awumey2024, Chandhiramowuli2024}, stakeholders should match task complexity to an annotator's measured cognitive profile.\cite{Heddaya2023} Aligning tasks with markers like working memory and inhibition control has been shown to increase accuracy by 5\% to 20\% while maintaining worker autonomy \cite{Hettiachchi2020}. This transition improves epistemic justice by replacing coercive monitoring with intellectual dignity.

\item Compensation models should account for the 33\% "invisible tax," representing the unpaid time workers spend on platform logistics and task search labor \cite{Toxtli2021}. Shifting from piecework to time-based or cooperative reward structures incentivizes the care and deliberation necessary for high-fidelity annotation. This prevents the impact of extreme time pressure, which typically forces workers into providing "safe" or generic answers that result in the systemic erasure of minority viewpoints \cite{Fan2020, Agarwal2020}.

\item To prevent the codification of algorithmic monocultures, along with better documentation, practitioners should institutionalize "human-anchor" protocols that resist the uncritical adoption of model-as-a-judge frameworks. As automated annotation scales, the reliance on superficial spot-checks facilitates a normative feedback loop where models recursively reinforce dominant linguistic patterns and filter out the "tail" distributions of diverse human thought \cite{shumailov2024ai, Kapania2025, blodgett2020, Felkner2024}. We recommend establishing situated adjudication layers where community-specific experts retain final veto power over automated syntheses, ensuring the human-in-the-loop remains a genuine conduit for lived experience.
\end{itemize}

\subsubsection{For Annotators: from data laborers to knowledge partners}

As the primary producers of "truth," annotators must be empowered as active stakeholders with the agency to shape the epistemological boundaries of the data they generate.

\begin{itemize} \item Annotators should be granted the right of refusal for projects that conflict with community values or exploit situated knowledge. To protect epistemic autonomy, platforms should implement structural safeguards to ensure that such refusals do not result in rating penalties or the systemic loss of future work opportunities \cite{DiSalvo2024, Ma2022}. This transition moves the domain away from extractive labor models toward a framework of mutual respect and ethical participation.

\item Researchers should move toward a co-design model where annotators collaborate on the creation of guidelines rather than functioning as passive recipients of external instructions. By documenting "discretionary calls" and interpretive rationales in update logs, the currently invisible labor of navigating data ambiguity becomes a legible asset for future audits. This practice transforms sense-making from a technical hurdle into a transparent component of the knowledge production process \cite{Miceli2022documenting, Miceli2021}.

\item In minoritized language contexts, the field should prioritize oral and narrative elicitation channels to bypass the inherent biases of Western, text-based interfaces. This shift is essential for preserving the subjectivity of spoken dialects and cultural metaphors that standardized, categorical surveys often strip away \cite{Reitmaier2024, Skeggs2025}. By adopting diverse communicative modes, the domain can better capture the richness of non-WEIRD knowledge systems.

\item By integrating frameworks such as Social Identity Mapping \cite{Jacobson2019} and Experience-Centered AI \cite{Gautam2025} into the annotator-to-task assignment layer, the pipeline can practically surface positionality-driven lenses.
\end{itemize}

\subsubsection{For Platforms and Developers: infrastructural inclusion}

Platform developers must address the structural filters that currently exclude the very populations required to decolonize training data and ensure global representativeness.

\begin{itemize} \item Where applicable, developers should prioritize mobile-accessible task design to integrate contributors from the Global South. Currently, the impact of desktop-first requirements is the effective silencing of 60\% of potential contributors in emerging economies, which reinforces a Western-centric "ground truth" \cite{Dutta2022}. Future epistemic justice depends on dismantling these technical barriers to ensure that the infrastructure of data production is as global as the AI systems it informs.

\item Platforms should transition from majority-vote models toward perspectivist adjudication systems that preserve the distribution of disagreement. Given that changing the demographic composition of an annotation jury can significantly alter model outcomes, maintaining these variations is essential for building systems that reflect pluralistic realities rather than a forced demographic average \cite{Gordon2022, LoraAroyo2023, Liu2022}. Moving toward latent-truth modeling, rationale extraction, or label distributions represents a fairer step toward more representative AI \cite{Cachel2025}.

\item To dismantle the myth of "hard-to-reach" populations, platforms should design interfaces that integrate with existing informal community channels, such as WhatsApp groups or localized guilds. These networks currently facilitate higher levels of engagement and trust in non-WEIRD contexts compared to isolated, centralized platforms \cite{De2025, Wang2023, Stureborg2023}. Leveraging these indigenous digital infrastructures allows for a more authentic and inclusive data collection process.
\end{itemize}

\subsection{Cross-Conference Trends in Data Annotation Research}
The ground truth illusion is refracted differently across disciplines, with each venue foregrounding a distinct stage of the annotation pipeline. Human-computer interaction venues, primarily CHI and CSCW, frame the problem as a failure of labor infrastructure and interface design, arguing that platform architectures force workers to prioritize throughput over perception—casting the consensus trap as a byproduct of constrained worker agency. FAccT and AIES treat it as a political crisis of epistemic justice, critiquing identity reductionism in classification schemas. NeurIPS focuses on technical debt and data cascades, where undocumented annotation choices propagate into downstream failures; this stream tends to treat human disagreement as a latent variable to be cleaned rather than a signal to be preserved. In natural language processing, ACL research emphasizes the interpretive gap in communication, investigating rationale extraction to mitigate anchoring bias and noting that linguistic meaning is often too fluid for static gold standards. EAAMO attends to geographic hegemony, critiquing Western recruitment norms and advocating for decolonized pipelines that integrate informal local channels to reach populations routinely dismissed as hard to reach.

A critical synthesis across venues reveals a structural divide: technical streams supply mechanisms for calculating agreement, but solutions for preserving the human voice consistently emerge from sociology and the humanities. Technical approaches tend to treat the "human-in-the-loop" as a sensor to be optimized for model convergence, assuming that with sufficient data or better mathematics, a singular truth can be recovered. Our findings show the opposite: purely mathematical aggregation produces a "manufactured cleanliness" that erases the very nuance required for social safety.
The sociological perspective reframes this entirely. Data is a situated artifact, not a raw resource, and the shift from extraction to stewardship requires interpretive methodology that technical streams alone cannot supply. For subjective domains (safety, toxicity, identity etc.) the "correct" label is not a statistical average but a distribution of lived realities. The future of the field therefore depends on treating data work not as an invisible technical nuisance but as a site of social negotiation, where the goal is not to eliminate disagreement but to make it legible.

\section{Limitations}

This review covers seven major venues from 2020--2025 (ACL, AIES, CHI, CSCW, EAAMO,
FAccT, NeurIPS), selected for their complementary coverage of annotation methodology
across ML, NLP, and socio-technical AI (Appendix~\ref{app:venues}). This scoping entails deliberate
trade-offs: by prioritizing premier Anglophone venues we underrepresent regional
conferences, workshop proceedings, and practitioner literature where community-led
practices are increasingly documented. We also exclude journals and non-peer-reviewed
sources. These constraints were necessary given the volume of publications over the
five-year window; broader venue coverage, including non-English and practitioner sources,
remains a priority for future work.

\section{Conclusion}

This systematic literature review deconstructs the positivistic fallacy of the "ground truth" paradigm, dissecting it as a socio-technical artifact manufactured through a sequence of deliberate and often invisible infrastructural choices. Our findings identify a systemic "consensus trap" where the prioritization of procedural standardization (\textit{niti}) over substantive justice (\textit{nyaya}) flattens cultural pluralism into statistical noise. We argue that current practices enforce a "manufactured cleanliness", a filtered reality driven by geographic hegemony and labor precarity that rewards sanitized objectivity over authentic subjectivity. To address these tensions, we propose a roadmap toward pluralistic infrastructures that prioritize situated knowledge stewardship and redefine expertise through the lens of lived experience. By institutionalizing perspectivist adjudication and reclaiming disagreement as a high-fidelity signal, the field can move beyond extractive data labor toward building culturally competent models grounded in human diversity.

\newpage
\section*{Acknowledgments}
This work was supported by the Natural Sciences and Engineering Research Council of Canada (NSERC), Google Award for Inclusion Research, and Connaught Fellowship for Community Partnership. The authors thank Zulkarin Jahangir (MIT Global Humanities
Initiative; UNESCO) for his contributions to this work.

\section*{Generative AI Disclosure}
In adherence to the ACM Policy on Authorship and the specific submission guidelines for FAccT 2026, the authors disclose the use of Large Language Models (LLMs) in the preparation of this manuscript. Google Gemini 3 Pro was utilized, exclusively for the purposes of improving grammatical fluency and assisting with the technical formatting of LaTeX tables. No original research text, conceptual frameworks, or empirical analyses were generated by AI. The authors have critically reviewed and edited all AI-assisted output and maintain full responsibility for the integrity and accuracy of the final work.


\bibliographystyle{ACM-Reference-Format}
\bibliography{references}

\section*{Appendix}
\begin{appendix}
\section{Venue-Year Coverage, Filtering Criteria, and RQ Alignment}
\label{app:venues}
\paragraph{Venue year coverage.}
Table~\ref{tab:venue_years} lists the publication years included for each target venue.

\begin{table}[h]
\caption{Publication years covered for each venue.}
\label{tab:venue_years}
\begin{tabular}{l c}
\toprule
\textbf{Venue} & \textbf{Years included} \\
\midrule
ACL     & 2020--2025 \\
AIES    & 2020--2024 \\
CHI     & 2020--2025 \\
CSCW    & 2020--2025 \\
EAAMO   & 2021--2025 \\
FAccT   & 2020--2025 \\
NeurIPS & 2020--2024 \\
\bottomrule
\end{tabular}
\end{table}

EAAMO was first held in 2021 and is therefore included from 2021 onward. FAccT 2020 was published under the former FAT* conference name and is included for continuity.

\paragraph{NeurIPS-specific filtering.}
Because NeurIPS contains a large volume of benchmark- and model-centric work where annotation is often mentioned only incidentally, NeurIPS records were processed with a stricter pre-screening procedure. We first scraped NeurIPS paper titles and abstracts from the official NeurIPS virtual proceedings pages (main conference papers only) and applied keyword matching over the concatenated title+abstract. We then applied a three-tier rule (Tier 1 sufficient signals; Tier 2 conditional signals requiring Tier 3 context anchors), but with a \emph{tighter context list} than the general pipeline to reduce false positives (e.g., matches such as ``ambiguity in dataset''). In particular, NeurIPS context anchors excluded generic data-centric terms (e.g., \textit{data}, \textit{dataset}) and instead required explicit annotation-related context (e.g., \textit{label}, \textit{annotation}, \textit{instruction}, \textit{worker}, \textit{rater}, \textit{judge}, \textit{human}).

\paragraph{Venues highlighting annotator selection and task-annotator fit (RQ1).} Human-centered and socio-technical venues such as ACM FAccT, AAAI/ACM AIES, ACM EAAMO, ACM CHI, and ACM CSCW frequently examine who performs annotation work and under what conditions. As such, they are particularly well-suited for addressing RQ1, which concerns how annotators are chosen, what metadata about annotators are collected, and where gaps remain in matching annotators to task demands.

\paragraph{Venues highlighting aggregation, disagreement, and formal annotation frameworks (RQ2).} Machine learning venues, including NeurIPS and ACL, frequently examine annotation through formal modeling lenses. As such, they are particularly well-suited for addressing RQ2, which concerns how multiple perspectives are represented, aggregated, preserved, or erased, how disagreement is quantified or reframed, and how collected reasoning is used or discarded during training and evaluation.

\section{Full-Text Exclusion Reasons}
\label{app:exclusions}
During full-text screening, 143 papers were excluded after failing to meet the eligibility criteria. Table~\ref{tab:exclusion_reasons} summarizes the primary reasons for exclusion and their respective counts.

\begin{table}[h]
\caption{Reasons for exclusion at the full-text screening.}
\label{tab:exclusion_reasons}
\begin{tabularx}{\linewidth}{@{} X r @{}}
\toprule
\textbf{Exclusion reason} & \textbf{Count} \\
\midrule
No annotation or labeling component & 59 \\
Wrong human role (end-user rather than annotator) & 26 \\
Standard benchmark dataset only (no new labeling or re-annotation) & 19 \\
Wrong publication type (e.g., abstract only) & 13 \\
Wrong focus: model performance metrics only & 12 \\
Wrong study design: secondary review & 10 \\
No human involvement in the pipeline & 2 \\
Wrong focus: model architecture only & 2 \\
\midrule
\textbf{Total} & 143 \\
\bottomrule
\end{tabularx}
\end{table}

\section{Full Keyword List for Tiered Pre-Screening}
\label{app:keywords}

This appendix reports the complete set of keywords used in the automated pre-screening stage.

\subsection{Tier 1: Sufficient Signals}
\begin{multicols}{2}
\begin{itemize}
  \item annotat*
  \item labell?ing
  \item ground truth
  \item gold standard
  \item reference standard
  \item human-generated
  \item dataset
  \item crowdwork
  \item crowdsourc*
  \item rater
  \item coder
  \item human-in-the-loop
  \item SME
  \item data work
  \item invisible work
  \item label noise
  \item noisy label
  \item inter-annotator
  \item inter-rater
  \item IAA
  \item Dawid--Skene
  \item GLAD
  \item MACE
  \item weak supervision
  \item RLHF
  \item human feedback
  \item preference optimization
  \item instruction tuning
  \item judging
  \item auto-evaluat*
  \item llm-as-a-judge
  \item generative annotation
  \item synthetic data
  \item persona-based
  \item model-as-annotator
  \item positionality
  \item epistemic justice
  \item perspectivis*
  \item soft label
  \item distributional label
  \item probabilistic label
  \item rating indeterminacy
\end{itemize}
\end{multicols}

\subsection{Tier 2: Conditional Signals}
\begin{multicols}{2}
\begin{itemize}
  \item aggregation
  \item consensus
  \item adjudication
  \item reconciliation
  \item rationale
  \item justification
  \item ambiguity
  \item subjectiv*
  \item disagreement
  \item bias
  \item multi-perspective
  \item pluralis*
  \item wage
  \item labor
  \item alignment
  \item simulat*
\end{itemize}
\end{multicols}

\subsection{Tier 3: Context Anchors}
\begin{multicols}{2}
\begin{itemize}
  \item label
  \item tag
  \item human
  \item worker
  \item rater
  \item judge
  \item annotation
  \item instruction
  \item dataset
\end{itemize}
\end{multicols}

\section{Geographical Breakdown}
\label{app:geographical_focus}

\begin{table}[H]
\centering
\small
\caption{Geographic focus of studies in the reviewed corpus ($N=346$). Global North contexts dominate the literature.}
\label{tab:geographic_focus}
\begin{tabular}{lrr}
\toprule
\textbf{Geographic Focus} & \textbf{Count} & \textbf{\%} \\
\midrule
Global North / WEIRD & 252 & 72.8 \\
Global South / Non-WEIRD & 35 & 10.1 \\
Mixed North \& South & 33 & 9.5 \\
Not Specified & 16 & 4.6 \\
Regional (East Asian etc.) & 10 & 2.9 \\
\midrule
\textbf{Total} & 346 & 100 \\
\bottomrule
\end{tabular}
\end{table}

\section{Task Domain Breakdown}
\label{app:task_breakdown}

\begin{table}[h]
\small
\caption{Primary annotation task domains in the reviewed corpus ($N=346$). Papers may
span multiple domains; counts reflect primary classification.}
\label{tab:tasks}
\begin{tabular}{lrr}
\toprule
\textbf{Task Domain} & \textbf{Papers} & \textbf{\%} \\
\midrule
Subjective Reasoning (toxicity, hate speech, sentiment, safety) & 99 & 28.6 \\
Domain-Specific Applied (medical, legal, education, science) & 52 & 15.0 \\
Miscellaneous / Cross-domain & 82 & 23.7 \\
Computer Vision (image tagging, segmentation, multimodal) & 48 & 13.9 \\
NLP (summarization, translation, benchmarking) & 36 & 10.4 \\
AI Alignment / RLHF & 29 & 8.4 \\
\midrule
\textbf{Total} & \textbf{346} & \textbf{100} \\
\bottomrule
\end{tabular}
\end{table}

\section{Taxonomy of Pre- and Post-Annotation Decisions in AI Pipelines}
\label{app:taxonomy}

\newcolumntype{L}{>{\raggedright\arraybackslash}p{0.45\textwidth}}
\newcolumntype{X}{>{\raggedright\arraybackslash}X}

\begin{center}
\onecolumn
\normalsize
\begin{xltabular}{\textwidth}{L X}
\toprule
\textbf{Stages} & \textbf{Supporting Papers} \\ \midrule
\endfirsthead

\multicolumn{2}{c}%
{{\bfseries \tablename\ \thetable{} -- Continued from previous page}} \\
\toprule
\textbf{Stages} & \textbf{Supporting Papers} \\ \midrule
\endhead

\midrule
\multicolumn{2}{r}{{Continued on next page}} \\
\bottomrule
\endfoot

\bottomrule
\endlastfoot

\noalign{\smallskip}\hline\noalign{\smallskip}
\multicolumn{2}{l}{\textit{\textbf{Pre-Annotation}}} \\
\noalign{\smallskip}\hline\noalign{\smallskip}

Annotator positionality as an architectural variable & \cite{Draws2022, Lerner2024, FloresSaviaga2023, Park2021-2, Liu2022, Noe2024, Oprea2020, Bhuiyan2020, Xie2022, Goyal2022, Ding2022, Kaufman2022, Scheuerman2025, CandiceSchumann2023, Belay2025, Casola2024, Yoo2024, Sachdeva2022, Jakesch2022, Sengupta2023, Wu2023, Rifat2024, Hall2024, Matsubara2021, Oprea2020, Phutane2025, Ovalle2023, Lameiro2025, Angerbauer2022, Mehta2025, Birhane2022} \\ \addlinespace

Recursive devaluation via synthetic data loops & \cite{Kapania2025, Wan2024, Norhashim2025, Whitney2024, AlvaradoGarcia2025, Phutane2025,  Magomere2025, Mehta2025, Buyl2025, Yaghini2021, He2025, Yeh2025, YannDubois2023, Hu2024, Leng2025, LuiseGe2024, HaotianQian2024, Fazelpour2025, Honovich2023, Beck2021, Arzaghi2025, Laskar2025, Calderon2025,  Miranda2025} \\ \addlinespace

The "Human-as-Verifier" and the validation bottleneck & \cite{YiqiJiang2024, Liu2023, Arzberger2025, Klinova2021, Lee2022, Li2024, He2025prompt, ChongyuQu2023, Cook2023, Ashktorab2021, Levy2021,Chu2025, Zagalsky2021, Schafer2025, Ji2025, AllenNie2023, Tolmeijer2022, Munyaka2023,  Xu2023, Park2025, Papenmeier2022, Min2025, Kapania2025, Cabrera2023, Ferguson2024, Huang2024b} \\ \addlinespace

Identity reductionism and the data feminism gap & \cite{Abdu2023, Barrett2023, Mickel2024, Bergman2023, Kambhatla2022, Khan2021, Cao2020, Boggust2025, Offenwanger2021, Suresh2022, Jourdan2025, Mayeda2025, Sengupta2023, Ovalle2023, Hettiachchi20202} \\ \addlinespace

System-level decisions as implicit annotators & \cite{Hall2024, Zajkac2023, Schumann2021, Fruchard2023, Hong2024, Anjum2021, Zhang2024, Jourdan2025, Magomere2025, Naggita2023} \\ \addlinespace

Defining expertise for annotation & \cite{Lameiro2025, Phutane2025, Davani2024, Uzor2021, Hall2024, Papakyriakopoulos2023, Harris2022, Proebsting2025, Dorn2024, WenYi2025, Jain2021, Reza2025, cCetin2021, Zhang2022-2, Verma2024, Fleisig2023, Chakrabarty2025, Mallari2020, Yang2021, Diaz2025, Jo2020, CruzBlandon2025, Kawakami2025, Bhardwaj2024, Sambasivan2022, Magdy2025, Valentine2024} \\ \addlinespace

Labor dynamics: invisible labor and performative alignment & \cite{Hawkins2023, Scheuerman2023, Wang2021, Kawakami2025, Kapania2025simulacrum, Wang2020, Steiger2021, Varanasi2022, Miceli2022, Maddalena2025, Agarwal2020, Simons2020, Oppenlaender2020, Jahanbakhsh2020, Ma2022, Boone2024, Tsvetkova2022, Awumey2024, Sannon2022, Arakawa2023, Kim2025, Li2025b} \\ \addlinespace

Geographic hegemony and the "hard-to-reach" myth & \cite{De2025, Arunkumar2025, Dingler2022, Reitmaier2024, Singh2025, Rifat2024, Pang2023, Abebe2021, InuwaDutse2023, Brown2025, Yang2024, Pistilli2025, Kak2020, simson2024, Lucchesi2022} \\ \addlinespace

Infrastructural filters and the "mobile penalty" & \cite{Randhawa2021, Hall2024, Pandey2020, Cambo2022, Dutta2022, Lucchesi2022, Munoz2022, Rostamzadeh2022} \\ \addlinespace

The capability vs. incentive paradox and variable effort & \cite{Chang2021, Hettiachchi2020, He2023, Wallace2025, Agarwal2020, Hettiachchi2021, Ali2021, Arzberger2025, Kraft2024, Miceli2022, Quaye2024, Lameiro2025} \\ \addlinespace

Community-centric participatory design as epistemic justice & \cite{Tan2024, Claisse2025, Lameiro2025, Li2025, Huang2022, AlvaradoGarcia2023, Jakesch2022, InuwaDutse2023, Zhang2025, Bondi2021, Rubya2021, Wang2022, Zhang2023, Parker2025, Jo2024, Gadiraju2023, Skeggs2025, Reitmaier2024, Lam2022, Rothschild2022, Dennler2023, Huang2024} \\

\noalign{\smallskip}\hline\noalign{\smallskip}
\multicolumn{2}{l}{\textit{\textbf{Post-Annotation}}} \\
\noalign{\smallskip}\hline\noalign{\smallskip}

The consensus trap: majority vote as a site of discrimination & \cite{Cachel2025, Gordon2021, Rizvi2025, LoraAroyo2023, Cachel2023, Ding2022, Wong2021, Trott2021, Johnson2021, MendezMendez2022, Santy2023, Chen2021, Burghardt2020, Schaekermann2020ambiguity, Behzad2025, Gordon2022, Haq2022, Lan2020, Lai2025, Jin2025, Luo2025} \\ \addlinespace

Pluralistic aggregation and perspectivist adjudication & \cite{ThornJakobsen2023, Weerasooriya2023, Davani2024, ThebaultSpieker2023, Leang2025, HadiHosseini2024, Gienapp2020, Fomicheva2020, Narimanzadeh2023, Xia2025, Xu2024, Casola2024, Gordon2022, Christoforou2021, Song2020, Peyrard2021, Zhao2025, Ueda2023, Wu2023, Beretta2021, Wu2024, SehyunHwang2023, Sun2025, LarsSchmarje2022} \\ \addlinespace

The "Noisy Sensor" fallacy and manufactured cleanliness & \cite{Rogers2021, Guerdan2023, Yang2020, Scheuerman2021, LeZhang2020, Chung2021,  Groh2022, JimenezSanchez2025, Thakkar2022, Hong2024, Brachman2022, Bi2023, Salim2024, Liu2024, Jinadu2024, Falk2025, BinHan2024, HuiGuo2024, Hanrahan2021} \\ \addlinespace

Deliberative annotation: from voting to reasoning & \cite{SchmerGalunder2025, Muller2021, Ortloff2023, Karadzhov2023, Schaekermann2020, Chen2024, Schaekermann2020ambiguity, Sharma2022, Abdelkadir2025, Kuo2024, Lai2023, Yao2023,  Mauri2024, Morrison2023, Yang2025, Wang2024, Ghai2021, Allen2022} \\ \addlinespace

Managerial bias in automated evaluation & \cite{Aoyagui2025, Phutane2025, Zheng2025, Kapania2025simulacrum, AlvaradoGarcia2025, Palla2025, Mok2023, Findeis2025, Sambasivan2021} \\ \addlinespace

Data cascades and documentation debt & \cite{Sambasivan2021, Ezema2025, Scheuerman2020, Darian2023, Fabris2022, Miceli2021, Hutchinson2021, Jo2020, Geiger2020, Rogers2023, Ramirez2021, JimenezSanchez2025, Muller2022, Anik2021, diaz2022, Boyd2021, Pushkarna2022, Heger2022, Chen2023, Poirier2025, Mishra2022, EmilyWenger2022, YanLiu2021, Oppenlaender2024, Arnold2025, Leavy2021} \\ \addlinespace

The snapshot fallacy and label decay & \cite{Randhawa2021, Heger2022, Bhardwaj2024, JimenezSanchez2025} \\ \addlinespace

Target-user mismatches and the expertise gap & \cite{Sauberli2024, Kornfield2022, Lin2024, Lai2022, Park2021} \\ \addlinespace

Post-hoc filtering as diversity-washing & \cite{Chan2024, Whitney2024, Nangia2021,  Fazelpour2025, Xue2021, Hong2024, Zhang2024} \\ \addlinespace

Metric hallucinations and the reliability gap & \cite{Gordon2021, Boag2021, Nangia2021, Sambasivan2021, Papenmeier2022} \\ \addlinespace

Operational barriers to pluralism & \cite{Kapania2023, Miceli2020, Heger2022, Birhane2022, Abdu2023, Miceli2021} \\ 
\end{xltabular}
\end{center}


\end{appendix}
\end{document}